\documentclass[pmlr]{jmlr}% new name PMLR (Proceedings of Machine Learning Research)

\usepackage{longtable}% for long tables

\usepackage{booktabs}
\usepackage[load-configurations=version-1]{siunitx} % newer version
\jmlrvolume{255}
\jmlryear{2024}
\jmlrworkshop{ML-DE Workshop at ECAI 2024}

\usepackage{acronym}
\acrodef{GAN} {Generative Adversarial Network}
\acrodef{ASR} {Automatic Speech Recognition}
\acrodef{UASR} {Unsupervised ASR}
\acrodef{DM}{Diffusion Model} 
\acrodef{TTS}{Text-To-Speech synthesis} 
\acrodef{EMD}{Earth Movers Distance}
\acrodef{PER}{Phone Error Rate}
\acrodef{WER}{Word Error Rate}
\acrodef{TDDM}{Trans-Dimensional Diffusion Model}
\acrodef{SDE}{Stochastic Differential Equation}
\acrodef{ODE}{Ordinary Differential Equation}
\acrodef{BPE}{Byte-Pair Encoding}
\acrodef{LM}{Language Model}
\acrodef{PP}{Perplexity}
\acrodef{MOS}{Mean Opinion Score}
\acrodef{MCD}{Mel-Cepstral Distortion}
\acrodef{DTW}{Dynamic Time Warping}
\acrodef{MSE}{Mean Square Error}
\acrodef{FID}{Fréchet Inception Distance}
\acrodef{CER}{Character Error Rate}
\acrodef{NODE}{Neural Ordinary Differential Equation}
\acrodef{NDE}{Neural Differential Equation}
\acrodef{TDNDE}{Trans-Dimensional Neural Differential Equation}
\acrodef{LL}{Log-Likelihood}
\acrodef{MR-VP}{Mean-Reverting Variance-Preserving}
\acrodef{VLB}{Variational Lower Bound}
\acrodef{CFG}{Classifier-Free Guidance}
\acrodef{SDXL}{Stable Diffusion XL}
\acrodef{BPD}{Bits-Per-Dimension}
\acrodef{TTI}{Text-To-Image}
\acrodef{AM}{Acoustic model}
\acrodef{NODE}{Neural \ac{ODE}}
\acrodef{}{}

\def\Exp{{\mathbb{E}}}

\def\param{{\mathbf{\theta}}}

\def\x{{\mathbf{X}}}

\usepackage{multirow}
\usepackage{pifont}
\title[What happens to diffusion model likelihood when your model is conditional?]{What happens to diffusion model likelihood when your model is conditional?}

  \author{\Name{Mattias Cross} \Email{mcross2@sheffield.ac.uk} \\ \and
   \Name{Anton Ragni} \Email{a.ragni@sheffield.ac.uk} \\
 \addr Speech and Hearing (SPandH), Dept. of Computer Science, The University of Sheffield, UK}

\editor{Editor's name}
\editors{C. Coelho, B. Zimmering, M. Fernanda P. Costa, L.L. Ferr\'as, O. Niggemann}

\begin{document}

\maketitle

\begin{abstract}
  %Introduction to field
%Generative models automatically produce data such as images, text and audio. It is also popular to use generative models to transform data of one modality to its equivalent context in another modality, as seen with \ac{TTI} AI art. % explain conditional
%
%Detailed background
\acp{DM} iteratively denoise random samples to produce high-quality data. %It is typical to use sampling methods derived from stochastic differential equations, allowing flexible inference. 
The iterative sampling process is derived from \acp{SDE}, allowing a speed-quality trade-off chosen at inference. Another advantage of sampling with differential equations is \emph{exact} likelihood computation. %These likelihoods have been used to evaluate unconditional \ac{DM}s, and likelihoods computed from \ac{DM}s have been applied to problems such as out-of-domain classification.
These likelihoods have been used to rank unconditional \ac{DM}s and for out-of-domain classification.
%
%General problem being addressed
Despite the many existing and possible uses of \ac{DM} likelihoods, the distinct properties captured are unknown, especially in conditional contexts such as \ac{TTI} or \ac{TTS}.
%
%Main result
Surprisingly, we find that \ac{TTS} \ac{DM} likelihoods are agnostic to the text input. \ac{TTI} likelihood is more expressive but cannot discern confounding prompts.
%
%What knowledge the main result reveals
Our results show that applying \ac{DM}s to conditional tasks reveals inconsistencies and strengthens claims that the properties of \ac{DM} likelihood are unknown.
%
%General context of results
This impact sheds light on the previously unknown nature of \ac{DM} likelihoods. Although conditional \ac{DM}s maximise likelihood, the likelihood in question is not as sensitive to the conditioning input as one expects. 
%
%Broader perspective
This investigation provides a new point-of-view on diffusion likelihoods.
\end{abstract}
\begin{keywords}
  Diffusion models, score-based generative modelling, likelihood
\end{keywords}

\section{Introduction} % (fold)
\label{sec:Introduction}
\acp{DM} learn to estimate a data distribution that can be sampled from, specifically through estimating the noise of an iterative denoising diffusion process. This training scheme can be seen as \ac{VLB} maximization \citep{hoDenoisingDiffusionProbabilistic2020,yangDiffusionModelsComprehensive2023}. A desirable feature of DMs is exact likelihood computation, where the likelihood of samples appearing in a data distribution can be calculated. Likelihood is predominantly used as an objective evaluation metric for sample quality. Although DMs are popular for text-guided synthesis, only unconditional tasks such as image-synthesis have focussed on likelihood. There is limited analysis on whether likelihood is a useful feature for conditional models, as used for \ac{TTI} and \ac{TTS}. Considering that DMs aim to maximize likelihood, learning what properties the likelihood has provides valuable research. This paper forms an initial survey on how likelihoods can be used in conditional scenarios and reviews any unexpected behaviour observed. We explore \ac{SDXL} \citep{podellSDXLImprovingLatent2023}, a \ac{TTI} model that uses \ac{CFG} \citep{hoClassifierFreeDiffusionGuidance2022} to generate images that depict text-prompts, and Grad-TTS \citep{popovGradTTSDiffusionProbabilistic2021}, a \ac{TTS} model that uses a \ac{MR-VP} \ac{SDE} to synthesise speech given a transcript. An acceptable assumption is that a conditional likelihood is a function that is sensitive to the input data. We test this assumption through a series of experiments on both models, and find that, for both conditioning mechanisms, there is a lack of insight as to how a conditional likelihood works. For example, SDXL likelihood cannot reliably pair images and captions. For Grad-TTS, we find likelihood is attributed voice quality and clean audio, but not inteligibility, leading to cases where audio sounds clean and the target speakers voice is correctly synthesised, but the speech is unintelligible.
%
%The rest of this review is organised as follows: A background on DMs, SDXL and Grad-TTS, a background and discussion on DM likelihood, experiments on both models, and concluding remarks.
% section Introduction (end)
% They are recognised for \ac{TTI} and \ac{TTS} applications, tasks known as conditional generation. 

\def\drift{{\mathbf{F}}}
\def\diff{{g}}
\def\d{{\mathrm{d}}}
\def\logdensity{{\nabla_\x\log p_t(\x)}}
\def\enc{{\mathbf{E}_\theta}}
\def\dec{{\mathbf{D}_\theta}}
\def\score{{\mathbf{S}_\theta}}
\def\clip{{\mathbf{T}_\theta}}
\def\pfode{{\mathbf{F}_\theta}}
\def\scalefactor{{\kappa(t)}}
\def\y{{\mathbf{y}}}
\def\x{{\mathbf{X}}}
\def\latent{{\mathbf{Z}}}
\def\prior{{\mathcal{N}(\mathbf{0}, \mathbf{I})}}
\def\w{{\mathbf{W}}}
\def\pfode{{\mathbf{H}_\param}}

\section{Background}

\subsection{Diffusion models}

\acp{DM} are usually viewed as denoisers. They are formulated as a Markovian process \citep{sohl-dicksteinDeepUnsupervisedLearning2015, hoDenoisingDiffusionProbabilistic2020}, or with \acp{SDE}~\citep{songScoreBasedGenerativeModeling2021}, as demonstrated in this study. %we refer to the latter in this work. 
A DM models a diffusion process from prior distribution $p_T(\x)$ to data distribution $p_0(\x)$. Given a forward noising process from $t=0$ to $t=T$ where $T$ is the terminal time-step, it is possible to transform a data sample $\x_0$ (e.g. an image or Mel-spectrogram) to a Gaussian sample $\x_T$ through a \ac{SDE} (Eq. \ref{eq:dm_forward}) with drift $\drift(\cdot)$ w.r.t. $t$ and diffusion $\diff(\cdot)$ w.r.t. the Wiener process $\w$.
\begin{equation}
    \d\x = \drift(\x, t) \d t + \diff(t)\d \w
    \label{eq:dm_forward}
\end{equation}
Eq. \ref{eq:dm_forward} is expressed by the following reverse-SDE when running backwards in time \citep{andersonReversetimeDiffusionEquation1982}
\begin{equation}
    \d\x = [\drift(\x, t) - \diff(t)^2\logdensity] \d t + \diff(t)\d \hat{\w}
    \label{eq:dm_backward}
\end{equation}
The gradient of the log-density $\logdensity$ is intractable, so it is estimated with a score-model $\score(\cdot)$, a neural network trained with score matching \citep{hyvarinenEstimationNonNormalizedStatistical2005, songSlicedScoreMatching2019, vincentConnectionScoreMatching2011}.
\cite{songScoreBasedGenerativeModeling2021} find that there is a deterministic counterpart to a reverse \ac{SDE} \ref{eq:dm_backward}, named a \textit{probability flow} \ac{ODE} (Eq. \ref{eq:dm_pfode})
\begin{subequations}
\begin{equation}
        \d\x = [\drift(\x, t) - \frac{1}{2}\diff(t)^2\logdensity] \d t
    \label{eq:dm_pfode}
\end{equation}
\begin{equation}
        \pfode(\x, t) = \drift(\x, t) - \frac{1}{2}\diff(t)^2\score(\x, t)
    \label{eq:dm_pfode_b}
\end{equation}
\end{subequations}
For inference, a trained score-model is used (Eq. \ref{eq:dm_pfode_b}).
The formulation of $\pfode(\x, t)$ is an example of a \ac{NODE} \citep{chenNeuralOrdinaryDifferential2019}. The connection with \acp{NODE} allows \textit{exact} likelihood computation of the ODE/SDE, which is the focus of this paper (Further addressed in Section~\ref{sec:ll_computation}).
Given that $p_T(\x_T)$ is similar to Gaussian noise, one can sample a Gaussian distribution and predict a real-sample $\x_0$ with a probability flow ODE (Eq. \ref{eq:dm_pfode_b}).

\subsection{Grad-TTS}

Grad-TTS \citep{popovGradTTSDiffusionProbabilistic2021} is a \ac{TTS} model that forms initial distributions centred on text encodings $\enc(\y)$ and applies a diffusion decoder to denoise into Mel-spectrograms. Grad-TTS uses a linear noise schedule (Eq. \ref{eq:beta_linear}) and an \ac{MR-VP}  SDE (Eq. \ref{eq:gtts_forward}) with ODE equivalent (Eq. \ref{eq:gtts_pfode}).
\begin{equation}
    \beta(t) = \beta_0 + (\beta_T - \beta_0)t
    \label{eq:beta_linear}
\end{equation}
\begin{equation}
    \d\x = \frac{1}{2}(\enc(\y) - \x)\beta(t) \d t + \sqrt{\beta(t)} \d\w
    \label{eq:gtts_forward}
\end{equation}
\begin{equation}
    \d\x = \frac{1}{2}((\enc(\y) - \x) - \score(\x, t, \enc(\y))) \beta(t) \d t
    \label{eq:gtts_pfode}
\end{equation}
The MR-VP SDE method used for modelling the conditional distribution $p_0(\x | \y)$ is different to CFG. Unlike CFG, there is no weighting parameter to control conditioning. We aim to clarify such differences in conditional methods. We experiment with this model as it is the first \ac{DM} for Mel-spectrogram decoding.

\subsection{Stable diffusion XL}

SDXL is a latent \ac{DM} for \ac{TTI} that boasts the highest Parti Prompts score.\footnote{At the time of writing, SDXL has a score of 33\%, 11\% higher than the next best model. \url{https://huggingface.co/spaces/OpenGenAI/parti-prompts-leaderboard}} It performs efficient sampling by processing images through a pretrained encoder and applying DM training within the latent space. The score estimator is a conditional U-Net \citep{ronnebergerUNetConvolutionalNetworks2015} that takes image latent $\x_t$ and OpenCLIP/CLIP text embeddings \citep{radfordLearningTransferableVisual2021, OpenCLIP}. In this work, we denote any network that encodes text $\y$ as $\enc(\cdot)$. Although SDXL was trained on multiple image sizes, sample quality is best for $1024\times 1024$. 
SDXL uses a sub-linear noise-schedule,
\begin{equation}
    \beta(t) = (\sqrt{\beta_0} + (\sqrt{\beta_T} - \sqrt{\beta_0})t)^2
\end{equation}
which is a time-dependant function that controls how much noise is added at each time-step. The cumulative perturbation of $\beta(t)$, known as denoising diffusion implicit sampling \citep{songDenoisingDiffusionImplicit2022}, is given by
\begin{equation}
    \sigma(t) = \sqrt{\frac{1 - \bar{\alpha}(t)}{\bar{\alpha}(t)}}
    \label{eq:ddpm_sigma}
\end{equation}
where $\bar{\alpha}(t) = \Pi_{s=0}^{t}\alpha(s)$ and $\alpha(t) = 1 - \beta(t)$. 
leading to the following ODE
\begin{equation}
\label{edm_light}
    \d\x =\score\left(\frac{\x}{\sqrt{\sigma(t)^2 + 1}}, t, \enc(\y)\right) \d\sigma(t)
\end{equation}
%
% and it's discretization
% %
% \begin{equation}
% \label{edm_discrete_light}
%     \x_{i+1} = \x_{i} + \score\left(\frac{\x_i}{\sqrt{\sigma_i^2 + 1}}, t_i, \enc(\y)\right)(\sigma_{i+1} - \sigma_i)
% \end{equation}
%
We refer readers to \cite{songDenoisingDiffusionImplicit2022, songScoreBasedGenerativeModeling2021} for the variance-exploding SDE that Eq. \ref{edm_light} is derived from. To generate images that depict text prompts, \ac{CFG} (Eq. \ref{eq:cfg}) \citep{hoClassifierFreeDiffusionGuidance2022} is used to condition model outputs on the text input by mixing the conditional ($\enc(\y)$) and unconditional ($\mathbf{0}$) model outputs 
\begin{equation}
    \label{eq:cfg}
    \score(\x, t, \enc(\y))^{(\omega)} = \score(\x, t, \mathbf{0}) + \omega[\score(\x, t, \enc(\y)) - \score(\x, t, \mathbf{0})]
\end{equation}
This introduces a \textit{guidance scale} parameter $\omega$ that controls the weight of the conditional output. Although increasing guidance improves sample accuracy, it trades sample quality. It is assumed that CFG models a conditional distribution $p_0(\x | \y)$, but the nuances of this assumption are clarified in this paper (Section~\ref{sec:tti}). We choose this model as it is the most accessible large \ac{TTI} \ac{DM} at the time of writing.

\def\d{{\mathrm{d}}}
\def\norm{{\mathcal{N}}}
\def\var{{\mathbf{1}}}
\def\drift{{\mathbf{f}}}
\def\diff{{g}}
\def\d{{\mathrm{d}}}
\def\scorekernel{{\nabla_{\x_t}\log p_{0t}(\x_t | \x_0)}}
\def\enc{{\mathbf{E}_\param}}
\def\real{{\mathbb{R}}}
\def\noisepred{{\nabla\x}}
\def\e{{\boldsymbol{\epsilon}}}
\def\grad{{d^2\x}}
% Likelihood estimation is a common task across many generative models e.g. VAE, NFLOW, HMM. There have been cases of conditional likelihoods for normalising flows for tasks such as super resolution \citep{winklerLearningLikelihoodsConditional2023}, but the sensitivity of the likelihood to the conditioning signal is not explored. GlowTTS \citep{kimGlowTTSGenerativeFlow2020} is a flow-based TTS model that maximises the conditional likelihood of speech given text. HMMs are well-known for conditional likelihoods for speech-to-text prediction.

\section{Diffusion likelihood computation}
\label{sec:ll_computation}
Albeit the lack of investigation into exact DM likelihoods, they are extensively applied in numerous contexts.\footnote{The use of ``exact'' is used to distinguish from lower and upper bounds, in practice we use an estimation of the ``exact'' likelihood.} Exact DM likelihood is computed with the instantaneous change-of-variables formula \citep{chenNeuralOrdinaryDifferential2019, grathwohlFFJORDFreeformContinuous2018}:
\begin{equation}
    \log p_0(\x_0) = \log p_T(\x_T) + \int^T_0 \nabla \cdot \pfode(\x, t) \d t
    \label{eq:diff_ll}
\end{equation}
Where $\pfode$ is a probability flow ODE (Eq. \ref{eq:dm_pfode_b}). Computing the exact divergence $\nabla \cdot \pfode(\x_t, t)$ is intractable, but can be estimated with the Skilling-Hutchinson trace estimator~\citep{skillingEigenvaluesMegadimensionalMatrices1989, hutchinsonStochasticEstimatorTrace1989}.
\begin{equation}
    \nabla \cdot \pfode(\x, t) = \Exp_{p(\e)}[\e^\top \nabla \pfode(\x, t)\e]
\end{equation}
where $\e$ is a random variable. The divergence integral $\int^T_0 \nabla \cdot \pfode(\x, t) \d t$ can be solved with a black-box ODE solver e.g. dopri5/RK45 \citep{dormandFamilyEmbeddedRungeKutta1980}. Many ODE solvers are efficiently implemented with \texttt{torchdiffeq} \cite{chenNeuralOrdinaryDifferential2019}. With a trained score-model, the likelihood $\log p_0(\x_0)$ is exact and can be used to measure the likelihood of a generated sample $\x_0$. 

% P1 unconditional
The relationship between likelihood and sample quality can be ambiguous~\citep{theisNoteEvaluationGenerative2016}. There is a trend for likelihood-based models that improve likelihood degrading other evaluation metrics e.g. \ac{FID}, without degrading visual quality \citep{songMaximumLikelihoodTraining2021}.  
Although important for evaluation, likelihood theory has alternatively been used for improved training and other tasks. \citeauthor{popovDiffusionBasedVoiceConversion2022} and \citeauthor{songMaximumLikelihoodTraining2021} use a likelihood-weighted upper bound to train models with better likelihood. \citeauthor{luMaximumLikelihoodTraining2022} estimate the likelihood to formulate high-order score matching, they stress the lack of understanding of $\log p_0(\x_0)$ yet produce good samples by maximising likelihood.
DM likelihoods can be applied to other tasks such as determining if a sample was used in training (membership inference \citep{huMembershipInferenceDiffusion2023}) and out-of-domain detection \citep{grahamDenoisingDiffusionModels2023}. 
 \citeauthor{popovGradTTSDiffusionProbabilistic2021} evaluate Grad-TTS with likelihood, but it is left unspecified whether this likelihood is conditional to the input text or not. 

% P2 conditional
The implications of using \textit{conditional} probability flow ODEs (Eq. \ref{eq:cfg}; \ref{eq:gtts_pfode}) are under-explored. This paper provides empirical confirmation on the current state of conditional DM likelihoods. Conditional likelihoods should correlate to high quality samples, but also compatibility between the generated sample and the conditioning signal. Although this assumption is intuitive, we demonstrate DM likelihood does not reflect compatibility as expected.

% P3 Misc
Diffusion classifiers show that \ac{VLB} performs well on conditional tasks. Image classification with \acp{DM} is alternative task that DMs have been applied to \citep{liYourDiffusionModel2023, clarkTexttoImageDiffusionModels2023}. Diffusion classifiers use \ac{VLB} as a proxy for likelihood to speed up computation. Although they have impractical inference time, they are useful for diagnosing DMs, such as textual robustness.
Nevertheless, there is no agreed definition for likelihood e.g. is it likelihood that a sample is from the training set? Is it the likelihood that a sample belongs to a given class? As there is no standard interpretation of what likelihood means, hence we explore the properties of likelihood.

\def\y{{\mathbf{y}}}
\def\enc{{\mathbf{E}_\mathbf{\theta}}}
\def\xblur{{\mathcal{N}(\x)}}
\newcommand{\xmark}{\ding{55}}
\section{Experiments}

% We provide a quantitative analysis both Grad-TTS and SDXL to observe how two different conditioning methods (\ac{MR-VP} \ac{SDE} and \ac{CFG}) affect likelihood on two different data types (audio and image). For both models we measure likelihood and other metrics during the backward-process to confirm that likelihood increases as $t$ decreases, and that likelihood correlates with other quality metrics. We test the limits of Grad-TTS likelihood on an n-best list rescoring experiment and on a domain-adaptation experiment. For SDXL we see how CFG scale affects likelihood. We find that processing an image through the forward-backward process can ``reconstruct'' the image such that likelihood is increased, without affecting other metrics. Finally, we test if SDXL likelihoods can accurately match visual-language captions from an array of confounds. For our experiments, we use the \texttt{torchdiffeq} package \citep{chenNeuralOrdinaryDifferential2019} to solve ODEs. 

\subsection{Text-to-speech}
The datasets we consider are the validation sets of LJSpeech \citep{LJSpeechDataset} and TED-LIUM \citep{rousseauTEDLIUMAutomaticSpeech2012}. The LJSpeech split contains around 1 hour of audiobooks read from a US-English female speaker. The TED-LIUM split contains around 3 hours of male speakers and 1 hour of female speaker TED conference audio.
To observe how likelihood changes during the denoising diffusion process, we generate the LJSpeech set and measure likelihood at 8 equally-spaced intervals of $t$, we expect that likelihood increases as $t$ decreases. We measure likelihood in \ac{BPD}, $-\log p(\x)\log_2\exp \cdot (\Pi_{i} \mathbf{d}_i)^{-1}$, where $d$ is the shape of $\x$. Lower BPD means higher likelihood.
Table \ref{tab:gtts_over_t} shows that although likelihood increases as $\x_T$ is denoised into data $\x_0$, the effect on other metrics is surprising. 
\begin{table}[htbp]
    \centering
    \caption{Quantitative assessment of Grad-TTS from time 1 to 0.}%Various metrics as time progresses from 1 to 0. These are likelihood (BPD), intonation error (LogF0), spectral distortion (MCD), intelligibility (WER), and speaker similarity (ASV). These results show that likelihood is correlated with smoother spectrograms and speaker accurate speech, but the diffusion generative process decreases intelligibility. There is a negligible change in intonation error.}
    \begin{tabular}{l||l|rrrrrrrrr}
        \toprule
        \multirow{2}*{Metric} & \multirow{2}*{Ref} & \multicolumn{9}{c}{Time}\\
         &  & 1.00 & 0.88 & 0.75 & 0.62 & 0.50 & 0.38 & 0.25 & 0.12 & 0.00 \\
        \midrule
        BPD $\downarrow$ & -0.06 & 2.79 & 2.79 & 2.79 & 2.79 & 2.77 & 2.72 & 2.55 & 1.35 & -0.08 \\
        LogF0 $\downarrow$ & 0.00 & 0.28 & 0.28 & 0.28 & 0.28 & 0.28 & 0.27 & 0.27 & 0.27 & 0.27 \\
        MCD $\downarrow$ & 0.00 & 6.52 & 6.52 & 6.52 & 6.50 & 6.47 & 6.43 & 6.39 & 6.34 & 6.31 \\
        WER $\downarrow$ & 0.03 & 0.08 & 0.08 & 0.08 & 0.09 & 0.08 & 0.10 & 0.13 & 0.16 & 0.19 \\
        ASV $\uparrow$ & 1.00 & 0.74 & 0.74 & 0.73 & 0.73 & 0.74 & 0.75 & 0.78 & 0.84 & 0.86 \\
        \bottomrule
    \end{tabular}
\label{tab:gtts_over_t}
\end{table}
% TODO could cite paper that says how whisper is correlated with intelligibility
\ac{MCD} is a measure of how different two Mel-cepstra (audio features) are from each other \citep{kominekSynthesizerVoiceQuality2008}.\footnote{We calculate MCD with the method presented in ESPnet \citep{gaoEUROESPnetUnsupervised2023}} We note the root mean squared error of the log fundamental frequencies as ``LogF0'', this measures pitch/intonation accuracy.\footnote{For reference, Tacotron 2 and Fastspeech 2 achieve 0.26 and 0.24 LogF0 respectively \citep{renFastSpeechFastHighQuality2022}.} \ac{WER} is the ratio of transcription errors to the number of words; to measure intelligibility we calculate the \ac{WER} of the transcript generated by the \ac{ASR} foundation model Whisper \citep{radfordRobustSpeechRecognition2022}. We measure speaker similarity (ASV) by comparing the cosine similarity between embeddings from the speaker verification foundation model Titanet \citep{koluguriTitaNetNeuralModel2021}.
Likelihood is correlated to ASV, marginally with MCD, but not LogF0. WER increases as likelihood improves. The results suggest that composition between the source text and hypothesis utterance\footnote{The ASR output} is modelled by the encoder $\enc(\cdot)$, and the diffusion decoder does not model the exact likelihood between text and speech $p_0(\x | \y)$. We verify this through an \ac{ASR} decoder rescoring experiment on LJSpeech. An ASR pipeline generates a list of hypothesis transcripts for a given utterance, which are scored by a weighted combination of an \ac{AM} score and a language model score to find the best transcription. We view this pipeline as a diagnostic tool for how well Grad-TTS likelihoods can measure speech/text compatibility. We replace the \ac{AM} score with Grad-TTS likelihoods and execute standard hypothesis rescoring techniques \citep{kahnFlashlightEnablingInnovation2022}. We test a wav2vec2 model finetuned on 10 minutes of audiobook data \citep{panayotovLibrispeechASRCorpus2015}, and Whisper (Table~\ref{tab:asr_reranking}).
\begin{table}[!htbp]
\centering
\caption{WER\% $\downarrow$ of n-best list rescoring strategies including Grad-TTS.}% Rescoring by DM likelihood is always worse than using the original \ac{AM}. This shows that the conditional likelihood $p_0(\x | \y)$ is not well expressed.}
\begin{tabular}{lSSSSS}
\toprule
\textbf{AM Model} & {AM Score} & {Grad-TTS LL} & {Oracle Best} & {Oracle worst} & {Random} \\
\midrule 
wav2vec2 & 13.3 & 19.6 & 10.5 & 29.0 & 20.2\\
Whisper & 3.3 & 4.7 & 1.5 & 7.2 & 4.3\\
\bottomrule
\end{tabular}
\label{tab:asr_reranking}
\end{table}
In all cases, the DM likelihoods yield a worse score than the \ac{AM}, although better than the oracle worst for both \acp{AM}.\footnote{Although Grad-TTS performs better than random for the wav2vec2 hypothesis list, this result is insignificant for an 82\% confidence interval with the matched-pairs significance test \citep{gillickStatisticalIssuesComparison1989}.} This suggests that the likelihoods may have some sensitivity to textual changes but lack an expressive linguistic representation. 
To survey how DM likelihood interacts with intelligibility and speaker quality, we employ Grad-TTS to adapt TED-LIUM live-talk data \citep{rousseauTEDLIUMAutomaticSpeech2012} to aid an ASR model trained on audiobooks. ASR models perform worse when the acoustic properties differ between train and test data. We pass the spectrogram through a Gaussian blur with kernel 5 and $\sigma=1$, simulating text encoder output $\xblur \approx \enc(\y)$.\footnote{This method is further explained in Appendix \ref{sec:Grad-TTS encoder and decoder output}} The forward ODE is used to calculate $\x_T$, $\x_0$ is produced with the reverse ODE thereafter. Given that Grad-TTS is trained on LJSpeech, the newly generated $\x_0$ will be acoustically similar to the data the ASR model was trained on. WER will be reduced if Grad-TTS can successfully adapt between live-talk and audiobook domains. We experiment with both Euler and RK45 solvers (Table~\ref{tab:uda}).
\begin{table}[!htbp]
    \centering
    \caption{Quantitative assessment of cross-domain adaptation with Grad-TTS.}% Results for adapting TED-LIUM livetalk data to LJSpeech audiobook style. The Euler sampler or RK45 is used to encode and decode speech. The Euler sampler produces speaker accurate but unintelligible data, there is an opposite trend when using RK45. Likelihood (BPD) has a similar correlation as seen in Table~\ref{tab:gtts_over_t}: more negative BPD means higher speaker similarity (ASV), but more unintelligible (WER). The choice of ODE sampler has a significant affect on speech adaptation.}
     \begin{tabular}{cccccc}
        \toprule
         \multicolumn{2}{c}{ODE sampler} & & \multicolumn{3}{c}{Metric}\\
         \cmidrule{1-2} \cmidrule{4-6}
         Forward & Reverse & & WER $\downarrow$ & BPD $\downarrow$ & ASV $\uparrow$ \\
         \midrule
          None & None & & 0.46 & 1.66 & 0.52\\
          Euler & Euler & & 0.94 & -1.47 & 0.66\\
          RK45 & Euler & & 0.57 & 6.92 & 0.54\\
          RK45 & RK45 & & 0.57 & 1.83 & 0.55\\
         \bottomrule
    \end{tabular}
    \label{tab:uda}
\end{table}
The Euler method produces unintelligible audio, yet has high speaker similarity and improved likelihood (BPD). The RK45 sampler degrades intelligibility to a lesser degree than Euler, but speaker adaptation is negligible. The trend is that greater shift towards the LJSpeech speaker (ASV increases) produces less intelligible audio (WER increases).

% The experiments above show that Grad-TTS likelihood is not correlated to the likelihood of observing a given speech sample in the data distribution, but the likelihood that a given sample ``sounds like'' it belongs to the data distribution. This has been shown through explicitly observing likelihood and other qualities through the generative process (Table~\ref{tab:gtts_over_t}). The likelihood was found to have below-standard utility on language understanding tasks (n-best list rescoring; Table~\ref{tab:asr_reranking}). Finally, the attributes of Grad-TTS likelihood are further established through a domain adaptation experiment (Table~\ref{tab:uda}) where speaker characteristics are improved but intelligibility is degraded, and the properties of likelihood measured is consistent with Table~\ref{tab:gtts_over_t}.

\subsection{Text-to-image}
\label{sec:tti}
\begin{figure}[htbp]
\floatconts
{fig:image_data}
{\caption{Example data from CLEVR ($a$) and PACS ($b$)}}% caption for whole figure
{%
\subfigure{%
\label{fig:red_sphere}% label for this sub-figure
\includegraphics[width=0.25\textwidth]{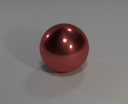}
} % space out the images a bit
\subfigure{%
\label{fig:pacs}% label for this sub-figure
\includegraphics[width=0.40\textwidth]{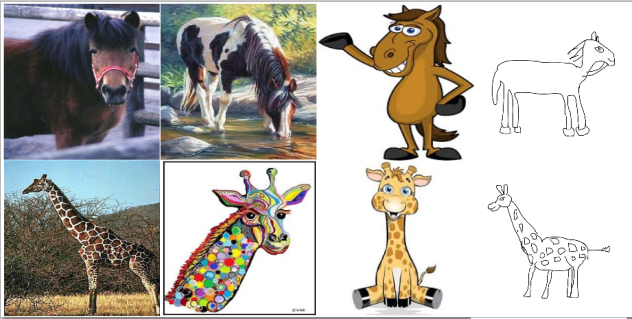}
}
}
\end{figure}

Datasets considered are PACS \citep{liDeeperBroaderArtier2017}, a domain generalisation dataset that includes images of 9 classes in 4 domains (photo, art, cartoon, sketch) and CLEVR \citep{johnsonCLEVRDiagnosticDataset2017}, a diagnostic dataset containing synthetic images of 3 3D objects of 8 colours. We use images generated from \cite{lewisDoesCLIPBind2023} e.g. a ``red sphere'' (Figure~\ref{fig:red_sphere}). We use a subset of CLEVR consisting of single objects. For both datasets we repeat each class-domain/shape-colour pair 20 times. Images are resized to $512^2$/$1024^2$.
We report the likelihood of SDXL, where we use \ac{CFG} to generate PACS. We compute \ac{FID} and CLIP similarity to measure image quality and compatibility respectively (Table~\ref{tab:sdxl_over_t}).
\begin{table}[!htbp]
\centering
\label{tab:sdxl_over_t}
\caption{The SDXL generative process from $t=1$ to 0 with CFG=7 at 1024 resolution.}
\begin{tabular}{l||l|rrrrrrr}
  \toprule
      \multirow{2}*{Metric} & \multirow{2}*{Data} & \multicolumn{5}{c}{Time}\\
                            & & 1.00 & 0.75 & 0.50 & 0.25 & 0.00 \\
  \midrule
      BPD $\downarrow$ & -0.17 & 0.07 & 0.08 & -0.08 & -0.12 & -0.08\\
      FID $\downarrow$ & 0.00 & 478.91 & 369.91 & 225.89 & 221.40 & 229.08\\
      CLIP $\uparrow$ & 29.50 & 22.72 & 23.49 & 28.10 & 27.97 & 28.16\\
  \bottomrule
\end{tabular}
\end{table}
FID measures the distance between real and synthetic image distributions. CLIP measures the compatibility between prompts and images. These metrics can be related to MCD and WER in the TTS experiment.  Similar to Table~\ref{tab:gtts_over_t}, FID and BPD both decrease, suggesting that the generative process improves sample quality and likelihood with respect to $t$. Although there is a clear trend that FID does decrease over time, it is higher than typical.\footnote{The DM in \cite{songScoreBasedGenerativeModeling2021} yields 2.92 FID on CIFAR-10.} It is assumed that this is because SDXL was trained on more data than PACS, thus generating many samples absent in the PACS dataset. Unlike Grad-TTS, the conditional metric (CLIP) improves, showing that CFG expresses a more conditional likelihood than the MR-VP SDE that used by Grad-TTS.
\begin{table}[!htbp]
  \centering
  \caption{Impacts of CFG, reconstruction (-R) and image sizes.}
  \label{tab:sdxl_cfg_reconstruct}
    \begin{tabular}{l||l|rrrr|rrrr}
      \toprule
          \multirow{2}*{Metric} & \multirow{2}*{Data} & \multicolumn{4}{c|}{CFG} & \multicolumn{4}{c}{Size-Reconstruct}\\
              & & 0.00 & 3.00 & 5.00 & 7.00 & 512 & 512-R & 1024 & 1024-R\\
      \midrule
          BPD $\downarrow$ & -0.17 & -0.04 & -0.05 & -0.07 & -0.08 & 0.02 & -0.01 & 0.07 & -0.01\\
          FID $\downarrow$ & 0.00 &  300.98 & 249.67 & 227.14 & 229.08 & 406.39 & 406.26 & 478.91 & 480.34\\
          CLIP $\uparrow$ & 29.50 & 23.16 & 26.70 & 27.68 & 28.16 & 22.73 & 22.75 & 22.72 & 22.71\\
      \bottomrule
    \end{tabular}
  \end{table}
Table~\ref{tab:sdxl_cfg_reconstruct} shows setting a higher CFG scale intuitively produces higher-likelihood samples. To measure how the generative process is intertwined with likelihood, we apply the forward process then backward process to an image producing an image similar to the original but with features more consistent to synthetic images. We dub this operation \textit{reconstruction}.\footnote{An example is given in Appendix \ref{sec:SDXL Images}} Reconstruction increases likelihood without affecting other metrics, revealing inconsistency (Table~\ref{tab:sdxl_cfg_reconstruct}).
Lastly, we investigated visual-language reasoning, an important ability for \ac{TTI} models. We are interested in observing whether DM likelihood can correctly match images to captions from a list of compositionally confounding captions. Although diffusion classifiers show that this is possible with proxy-likelihoods, can the same be said with exact likelihood? We evaluate the ability to discern objects on CLEVR. For each image, we present all confounding prompts e.g. the correct prompt is ``a photo of a red sphere'' but instead ``a photo of a red cube'' is provided. Accuracy is calculated based on the number of correct captions that were given the highest likelihood. We confound both colour and shape, but not at the same time. 
Correspondingly, we are interested in other visual-language concepts.  Complementary to the CLEVR experiment, we use PACS to measure the likelihood sensitivity to object vs domain changes e.g. a photo of a horse (Figure~\ref{fig:pacs}). 
\begin{table}[!htbp]
  \caption{Prompt accuracy on CLEVR and PACS}\label{tab:clevr_pacs}
  \centering
  \begin{tabular}{rlrrrrrr}
    \toprule
    \multicolumn{2}{c}{Params} && \multicolumn{2}{c}{CLEVR} && \multicolumn{2}{c}{PACS}\\
        \cmidrule{1-2} \cmidrule{4-5} \cmidrule{7-8}
    Reconstruct & CFG && Colour & Shape && Class & Domain\\
    \midrule
    \xmark & 0 && 12 & 33 && 86 & 50 \\
    & 3 &&  0 & 33 && 0 & 0 \\
    & 5 && 25 & 33 && 0 & 0 \\
    % & 7 && 12 & 33 && 0 & 0 \\
    \midrule
    \checkmark & 0 && 38 & 33 && 14 & 0 \\
    & 3 && 12 & 0 && 14 & 0 \\
    & 5 && 12 & 0 && 0  & 0 \\
    % & 7 && 12 & 0 && 0  & 0 \\
    \midrule
    Random &&& 12 & 33 && 11 & 25 \\
    CLIP-score &&& 63 & 67 && 100 & 100 \\
    \bottomrule
  \end{tabular}
\end{table}
The results are shown in Table \ref{tab:clevr_pacs}. It can be seen that the scores are unexpectedly low. We also provide scores from random selection and when selecting captions with the highest CLIP score. The results on CLEVR are comparable to random choice, when reconstruction is used, shape accuracy is reduced. SDXL performs better on PACS without reconstruction and guidance. This is anomalous behaviour since the rest of the PACS results are typically 0\% accuracy. 

% These experiments confirm that the sampling process in SDXL improves likelihood, image quality, and prompt faithfulness over time. The prompt faithfulness is correctly controlled with guidance scale. DM likelihood is ineffective for prompt classification. Such results show that the conditional modelling in SDXL produces inaccurate likelihoods. \citeauthor{clarkTexttoImageDiffusionModels2023} can perform shape and colour binding on the CLEVR dataset with the \ac{VLB} of the likelihood, but this is not the case when using the exact likelihood, exposing a quantitative gap in DM understanding.

\section{Conclusion}
Given that DMs are trained to maximise the \ac{VLB} of likelihood, it is important to understand the nuances of this exact likelihood. This is especially true for conditional tasks where the conditional likelihood is often modelled implicitly, such as with a \ac{MR-VP} SDE, or with \ac{CFG}. We show that any reasonable assumptions should by verified as the task and conditional ODE used has substantial impact on the nature of DM likelihood. For Grad-TTS, the \ac{MR-VP} SDE likelihood models speaker characteristics and smooth spectrograms, but not features of language. Such findings are contrary to what one would expect for a probabilistic conditional process between speech and text. For SDXL, CFG does appear to model prompt-faithful image generation, but exact likelihood from SDXL cannot be used for prompt classification. This is inconsistent with the concept of Diffusion Classifiers that use lower bounds on likelihood for classification. Hence, we quantitatively show a current lack of understanding of DM likelihood. This paper represents introductory evidence that more attention should be placed in experiments and theoretical understanding in DM likelihood, especially on conditional tasks.

\section*{Limitations}
This paper intends to demonstrate the potential adverse effects of introducing (implicit) conditioning mechanisms into frameworks originally studied for unconditional modelling. This paper does not provide any theoretical explanations to the unexpected behaviour observed, nor are any new explicit conditional likelihood methods derived. This paper has only explored two types of conitional generation, and only with their respective ``iconic'' model, whether this generalises to all diffusion models is unknown. Diffusion models are an instance of a continuous normalizing flow, a natural extension is to observe if other continuous normalizing flows e.g. flow-matching models exhibit the same behaviour.

\acks{We thank Peter Vickers for his helpful knowledge of the CLIP model, Kane O`Reagan \& Shaun Cassini for proof-reading, and Xiaozhou Tan for help with figures. This work was supported by the Centre for Doctoral Training in Speech and Language Technologies (SLT) and their Applications funded by UK Research and Innovation [grant number EP/S023062/1].}

\bibliography{./references.bib}

\newpage
\appendix
\section{Grad-TTS encoder and decoder output} % (fold)
\label{sec:Grad-TTS encoder and decoder output}
\begin{figure}[htbp]
\floatconts
{fig:gtts_encoder}
{\caption{Grad-TTS encoder output ($a$) and decoder output ($b$). The encoder produces intelligible spectrograms, and the decoder removes distortion and encourages speaker characteristics. The fact that the encoder output is similar to distorted spectrograms is core to the unsupervised domain adaptation method in Table \ref{tab:uda} where blurry spectrograms are treated as input to the diffusion decoder.}}% caption for whole figure
{%
\subfigure{%
\includegraphics[width=0.45\textwidth]{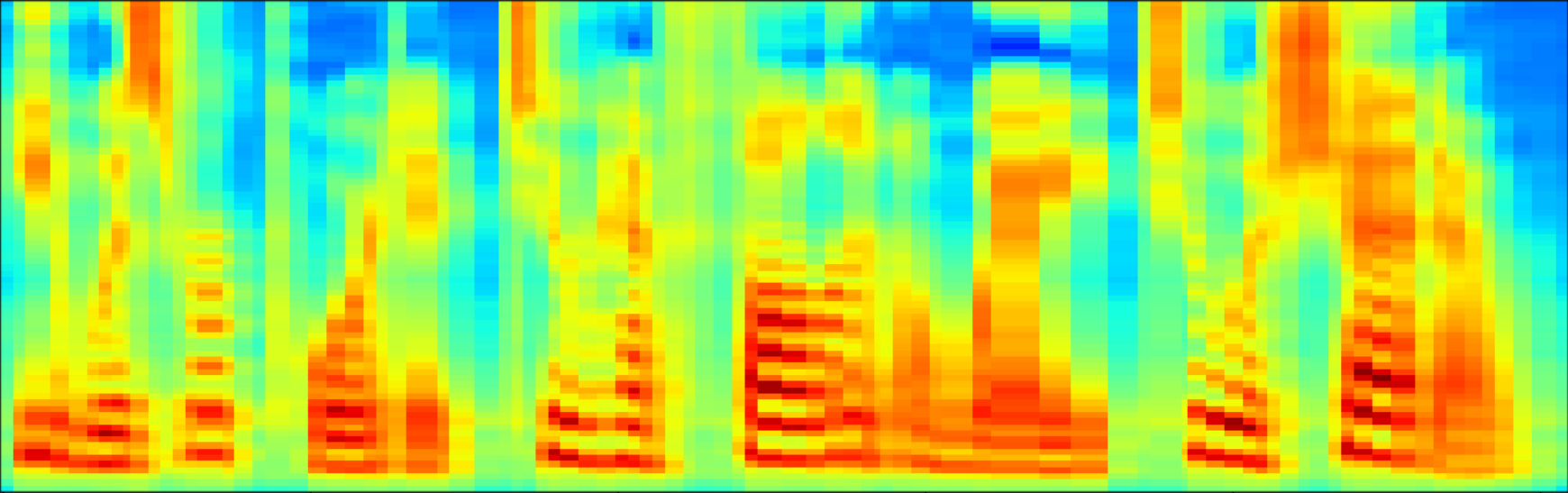}
}\qquad % space out the images a bit
\subfigure{%
\includegraphics[width=0.45\textwidth]{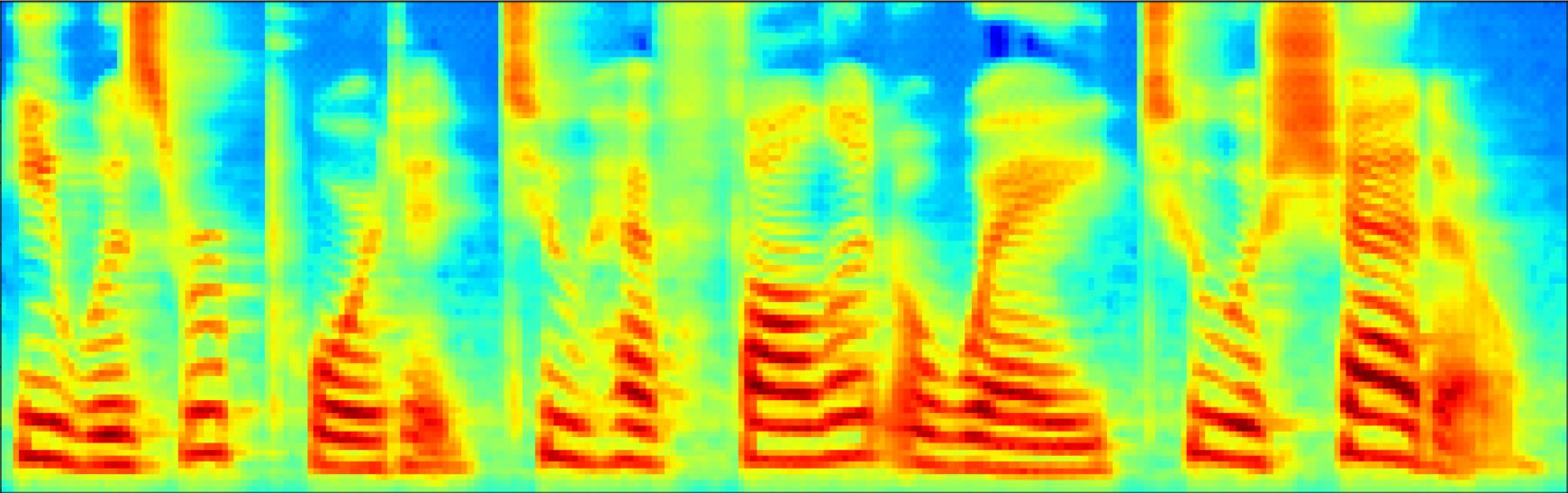}
}
}
\end{figure}
\newpage
% section Grad-TTS encoder and decoder output (end)
\section{Additional tables} % (fold)
\label{sec:tables}
% TODO fix big table
\begin{table}[ht]
  \caption{Various metrics during the generative process over time. The image distribution quality (FID) and prompt accuracy (CLIP) improve, as does the likelihood (BPD). The best results are found at 1024 resolution with a high guidance scale.}
\tiny
    \begin{tabular}{rrrrrrrrrrrrrrrrrrrrr}
\toprule
\multicolumn{3}{c}{pipe} & \multicolumn{3}{c}{1.00} & \multicolumn{3}{c}{0.75} & \multicolumn{3}{c}{0.50} & \multicolumn{3}{c}{0.25} & \multicolumn{3}{c}{0.00} & \multicolumn{3}{c}{ref} \\
CFG & height & r & BPD & FID & CLIP & BPD & FID & CLIP & BPD & FID & CLIP & BPD & FID & CLIP & BPD & FID & CLIP & BPD & FID & CLIP \\
\midrule
0 & 512 & \xmark & 0.02 & 417.58 & 22.58 & 0.02 & 399.30 & 23.62 & -0.01 & 310.19 & 22.68 & -0.01 & 308.96 & 22.23 & -0.01 & 310.24 & 22.55 & -0.04 & 0.00 & 29.56 \\
3 & 512 &  & 0.02 & 412.40 & 22.60 & 0.02 & 391.65 & 23.90 & -0.01 & 280.20 & 23.75 & -0.02 & 279.52 & 23.45 & -0.01 & 278.99 & 23.66 & -0.04 & 0.00 & 29.56 \\
5 & 512 &  & 0.02 & 409.33 & 22.67 & 0.02 & 378.74 & 24.18 & -0.02 & 268.82 & 24.36 & -0.02 & 268.72 & 23.95 & -0.02 & 264.01 & 24.35 & -0.04 & 0.00 & 29.56 \\
7 & 512 &  & 0.02 & 406.39 & 22.73 & 0.02 & 373.26 & 24.31 & -0.02 & 257.55 & 24.63 & -0.03 & 257.93 & 24.31 & -0.02 & 257.00 & 24.56 & -0.04 & 0.00 & 29.56 \\
0 & 512 & \checkmark & 0.01 & 417.80 & 22.58 & 0.00 & 399.60 & 23.61 & -0.01 & 309.89 & 22.65 & -0.02 & 308.74 & 22.23 & -0.01 & 311.61 & 22.57 & -0.04 & 0.00 & 29.56 \\
3 & 512 &  & -0.00 & 412.18 & 22.60 & -0.00 & 392.58 & 23.93 & -0.02 & 282.78 & 23.87 & -0.02 & 278.91 & 23.47 & -0.02 & 280.99 & 23.67 & -0.04 & 0.00 & 29.56 \\
5& 512 &  & -0.01 & 408.25 & 22.68 & -0.01 & 374.78 & 24.19 & -0.02 & 269.79 & 24.41 & -0.03 & 266.02 & 24.08 & -0.02 & 265.90 & 24.28 & -0.04 & 0.00 & 29.56 \\
7 & 512 &  & -0.01 & 406.26 & 22.75 & -0.01 & 373.70 & 24.31 & -0.02 & 256.47 & 24.63 & -0.03 & 256.94 & 24.36 & -0.02 & 259.19 & 24.66 & -0.04 & 0.00 & 29.56 \\
0 & 1024 & \xmark & 0.07 & 478.22 & 22.59 & 0.08 & 418.89 & 21.98 & -0.04 & 300.59 & 23.00 & -0.06 & 309.11 & 22.99 & -0.04 & 300.98 & 23.16 & -0.17 & 0.00 & 29.56 \\
3 & 1024 &  & 0.07 & 478.30 & 22.65 & 0.08 & 399.29 & 22.51 & -0.05 & 245.04 & 26.55 & -0.08 & 251.32 & 26.59 & -0.05 & 249.67 & 26.70 & -0.17 & 0.00 & 29.50 \\
5 & 1024 &  & 0.07 & 478.65 & 22.70 & 0.08 & 380.89 & 23.04 & -0.07 & 231.44 & 27.72 & -0.10 & 232.38 & 27.69 & -0.07 & 227.14 & 27.68 & -0.17 & 0.00 & 29.51 \\
7 & 1024 &  & 0.07 & 478.91 & 22.72 & 0.08 & 369.91 & 23.49 & -0.08 & 225.89 & 28.10 & -0.12 & 221.40 & 27.97 & -0.08 & 229.08 & 28.16 & -0.17 & 0.00 & 29.47 \\
0 & 1024 & \checkmark & 0.06 & 478.22 & 22.59 & 0.08 & 418.82 & 21.97 & -0.04 & 300.53 & 23.02 & -0.06 & 308.12 & 22.99 & -0.04 & 301.76 & 23.17 & -0.17 & 0.00 & 29.56 \\
3 & 1024 &  & 0.05 & 479.91 & 22.65 & 0.03 & 402.99 & 22.49 & -0.06 & 252.33 & 26.55 & -0.09 & 263.05 & 26.60 & -0.06 & 255.12 & 26.77 & -0.17 & 0.00 & 29.48 \\
5 & 1024 &  & 0.02 & 480.88 & 22.68 & 0.00 & 384.40 & 23.01 & -0.07 & 239.35 & 27.69 & -0.11 & 241.89 & 27.66 & -0.07 & 235.62 & 27.73 & -0.18 & 0.00 & 29.47 \\
7 & 1024 &  & -0.01 & 480.34 & 22.71 & -0.01 & 375.69 & 23.43 & -0.09 & 231.81 & 27.96 & -0.13 & 228.23 & 27.95 & -0.09 & 234.73 & 28.10 & -0.18 & 0.00 & 29.48 \\
\bottomrule
\end{tabular}
    \label{tab:sdxl_over_t_full}
\end{table}

\section{SDXL Images} % (fold)
\label{sec:SDXL Images}
\begin{figure}[htbp]
\floatconts
{fig:reconstruct}
{\caption{A source image ($a$) and a reconstructed image ($b$), with caption ``panda eating cake''. The panda and orientation are preserved but the domain has changed. The cake has been absorbed.}}% caption for whole figure
{%
\subfigure{%
\includegraphics[width=0.4\textwidth]{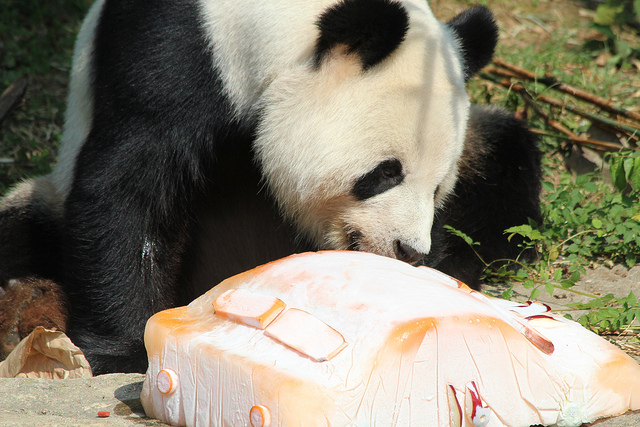}
}\qquad % space out the images a bit
\subfigure{%
\includegraphics[width=0.3\textwidth]{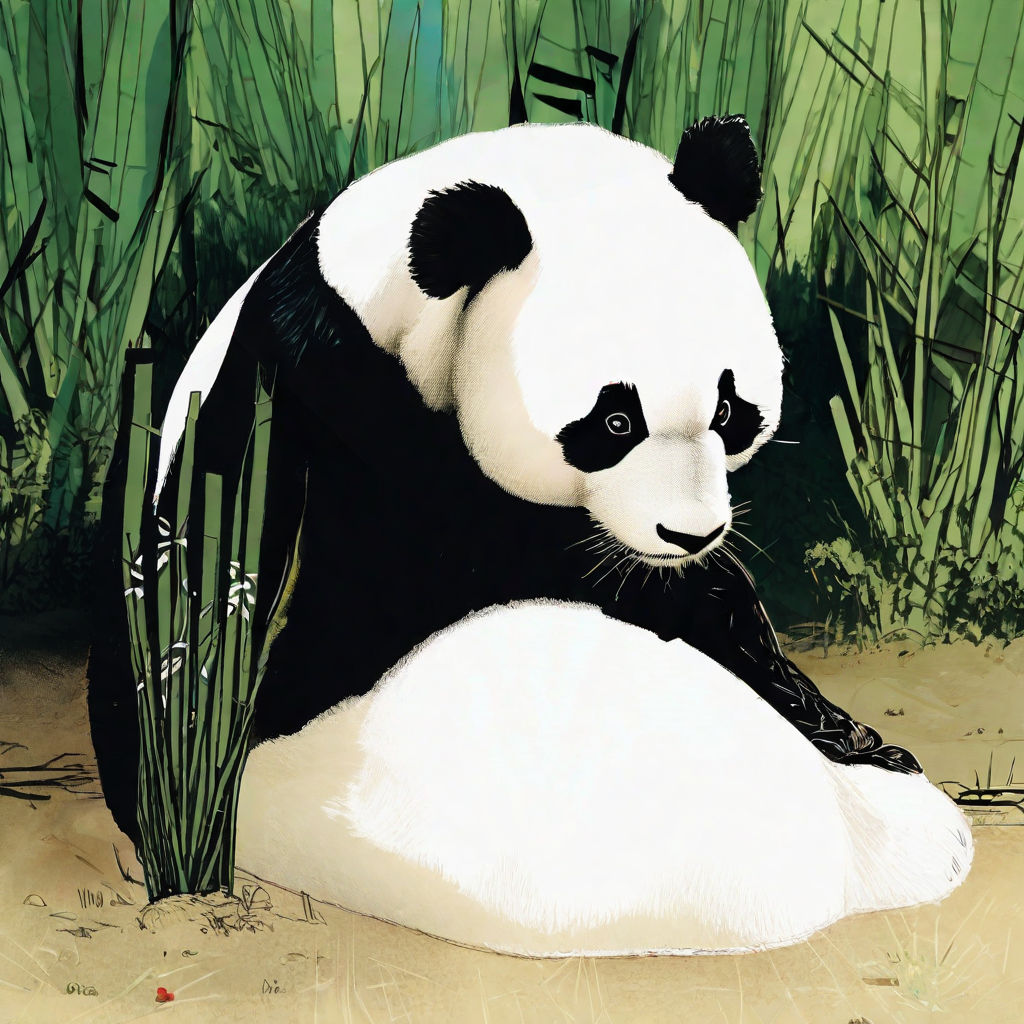}
}
}
\end{figure}

% section SDXL images (end)

\end{document}